\documentclass[letterpaper, 10 pt, conference]{ieeeconf} 

\IEEEoverridecommandlockouts          
                   \usepackage{graphicx}
\usepackage{amsmath}
\usepackage{amssymb}
\usepackage{booktabs}
\usepackage{multirow}
\usepackage{multicol}
\let\labelindent\relax
\usepackage{enumitem}
\usepackage{adjustbox}
\usepackage{cite}

\usepackage{color}

\title{\LARGE \bf
MetaFold: Language-Guided Multi-Category Garment Folding Framework via Trajectory Generation and Foundation Model
}

\author{Haonan Chen$^{*1,2}$, Junxiao Li$^{*3}$, Ruihai Wu$^{*4}$, Yiwei Liu$^{1}$, Yiwen Hou$^{1}$, Zhixuan Xu$^{1}$,\\ Jingxiang Guo$^{1}$, Chongkai Gao$^{1}$, Zhenyu Wei$^{5}$, Shensi Xu$^{3}$, Jiaqi Huang$^{3}$, Lin Shao$^{1,2\dagger}$% <-this % stops a space
\thanks{* denotes equal contribution}% <-this % stops a space
\thanks{$\dagger$ denotes the corresponding author}%
\thanks{$^{1}$Haonan Chen, Yiwei Liu, Yiwen Hou, Zhixuan Xu, Jingxiang Guo, Chongkai Gao, and Lin Shao are with the School of Computing, National University of Singapore
        }%
\thanks{$^{2}$Haonan Chen and Lin Shao are also with the NUS Guangzhou Research Translation and Innovation Institute
        }%
\thanks{$^{3}$Junxiao Li, Shensi Xu, and Jiaqi Huang are with the School of Computer Science, Nanjing University}
\thanks{$^{4}$Ruihai Wu is with the School of Computer Science, Peking University}
\thanks{$^{5}$Zhenyu Wei is with the School of Electronic Information and Electrical
Engineering, Shanghai Jiao Tong University}
}

\begin{document}

\maketitle
\thispagestyle{empty}
\pagestyle{empty}

%%%%%%%%%%%%%%%%%%%%%%%%%%%%%%%%%%%%%%%%%%%%%%%%%%%%%%%%%%%%%%%%%%%%%%%%%%%%%%%%
\begin{abstract}
Garment folding is a common yet challenging task in robotic manipulation. The deformability of garments leads to a vast state space and complex dynamics, which complicates precise and fine-grained manipulation. In this paper, we present \textbf{MetaFold}, a unified framework that disentangles task planning from action prediction and learns each independently to enhance model generalization. It employs language-guided point cloud trajectory generation for task planning and a low-level foundation model for action prediction. This structure facilitates multi-category learning, enabling the model to adapt flexibly to various user instructions and folding tasks. We also construct a large‑scale MetaFold dataset comprising folding point cloud trajectories for a total of 1210 garments across multiple categories, each paired with corresponding language annotations. Extensive experiments demonstrate the superiority of our proposed framework. Supplementary materials are available on our website: https://meta-fold.github.io/.
% Website: \href{https://meta-fold.github.io/}{https://meta-fold.github.io/}
\end{abstract}
\section{Introduction}
\label{sec:intro}

One of the major challenges in general robotic applications is the manipulation of deformable objects~\cite{zhu2022challenges}. Tasks that humans perform intuitively, such as folding clothes, present significant challenges for robots. This difficulty arises because deformable objects exhibit diverse forms of deformation, resulting in a large state space and complex dynamics~\cite{gu2023survey}. These factors complicate the prediction of future configurations, the planning of manipulation points, and the precise control of force and positioning.

Humans manipulate deformable objects effectively by separating high-level task planning from low-level control. The brain oversees task understanding and object recognition, while the spinal cord and peripheral nervous system manage hand movements and grasping. Inspired by this structure, we disentangle the complex dynamics of garment manipulation into future state prediction (planning) and low-level manipulation (execution), enabling a more modular approach to these challenging tasks.

\begin{figure}[t]
  \centering
  \includegraphics[width=1.0\linewidth]{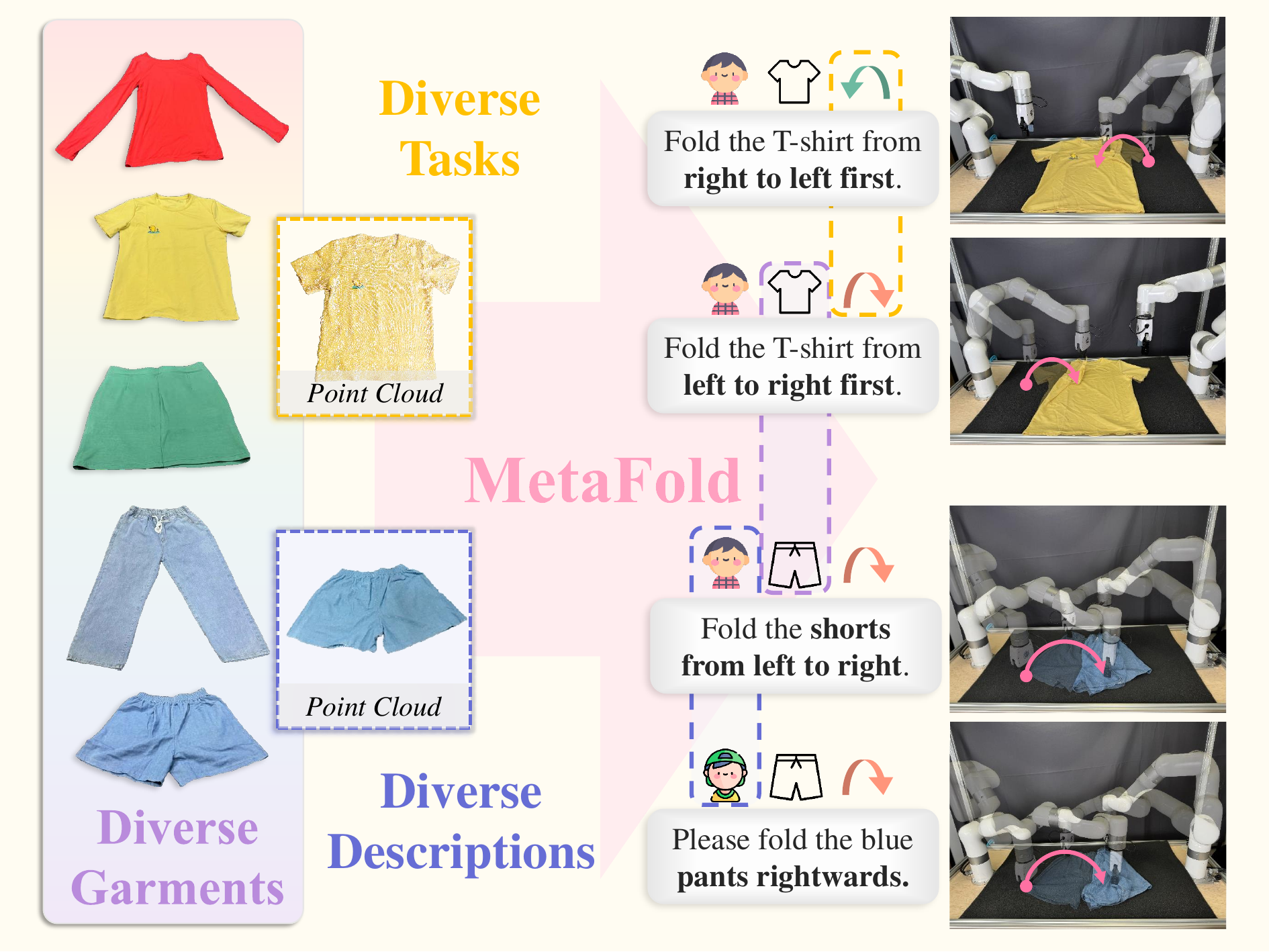}
  \vspace{-18pt}
  \caption{We present \textbf{MetaFold}, a unified framework capable of handling diverse garments and a wide range of language instructions, enabling various clothing folding tasks efficiently.} 
\label{fig:teaser}
\vspace{-15pt}
\end{figure}

Predicting future states and actions from arbitrary initial states introduces highly complex dynamics, posing significant challenges to the learning process. By contrast, decomposing the task and focusing specifically on predicting garment states during the folding process simplifies the learning process and further facilitates multi-category generalization, benefiting from the strong capabilities of generative models. 

This approach leverages the relatively simple and well-structured trajectories that characterize state transitions in garment folding, whereas modeling actions applied to arbitrary states would require capturing a broad range of implicit physical laws. 
Consequently, although accurately modeling full dynamics is challenging, predicting state transitions (i.e., a sequence of future states) for garment folding is comparatively feasible when action prediction is performed separately and conditioned on these transitions.

Accurate prediction of future states in garment manipulation requires an effective representation of garment configurations. Point cloud trajectories fulfill this need by providing a comprehensive spatial model that captures any point in space and encodes temporal changes in object states.
However, existing methods often rely on structural representations, such as key points or skeletal models~\cite{shi2021skeleton}, and depend on demonstrations or heuristic-based policies~\cite{ganapathi2021learning, wu2024unigarmentmanip, hietala2022learning, he2024fabricfolding, canberk2023cloth}. This dependence restricts these models to specific object configurations, limiting their adaptability across garment categories. In contrast, by using point cloud trajectories instead of structural representations, we maintain spatial and temporal awareness of the states, enabling more precise garment folding. Additionally, employing generative model to create point cloud trajectories facilitates multi-category learning and generalization within a unified framework, yielding robust performance across diverse garment manipulation tasks.

Additionally, point cloud trajectory generation facilitates generalization across both language inputs and tasks. Previous heuristic-based and demonstration-based methods struggle to perform different tasks under the same configuration. In contrast, our method leverages language-conditioned trajectory generation, enabling it to adapt dynamically to a wide range of user-specified instructions.

To translate planned garment trajectories during manipulation into executable actions for the robot, it is necessary to bridge trajectory planning with actionable outputs. While some foundational models that rely on keypoint detection can achieve basic folding tasks to a limited extent~\cite{huang2024rekep, raval2024gpt}, they often fail to account for intermediate garment states, leading to suboptimal outcomes.
The ManiFoundation model~\cite{xu2024manifoundation}, a general-purpose robotic manipulation module, addresses this limitation by transforming tasks into point flow and synthesizing contact points for low-level manipulation actions. By segmenting trajectories into multiple intermediate states, the ManiFoundation model effectively proposes actions between successive object states across various objects.
This capability allows us to disentangle robot action prediction from the overall manipulation procedure, thereby enabling a more focused approach to task understanding and trajectory generation. To enhance model robustness, we propose a closed-loop framework that integrates the point cloud trajectory generation model with the ManiFoundation model for action prediction. This framework also supports trajectory generation from intermediate states, allowing for adjustments based on the current folding state.

In summary, our main contributions are as follows:
\begin{itemize}
\item We introduce a novel framework for garment manipulation that disentangles state planning from action prediction. This hierarchical separation enables the model to focus on components, reducing training complexity.
\item We propose \textbf{MetaFold}, a framework that integrates a language-guided point cloud trajectory generation model with an action prediction foundation model, thereby facilitating multi-category garment folding.
\item We developed the point cloud trajectory dataset for folding garments across multiple categories, accompanied by corresponding language descriptions.
\item We conduct extensive experiments demonstrating MetaFold's superior performance in folding accuracy and language generalization.
\end{itemize}
\section{Related Work}
\label{sec:related}

\subsection{Garment Folding}

Garment folding, a challenging task owing to the deformable and high-dimensional nature of fabrics, has been extensively studied in the literature~\cite{ hoque2022learning, lee2021learning, tanaka2018emd, stria2014garment}.

Recent advancements have addressed these challenges through various approaches, including visual perception, dynamic control, and reinforcement learning. Several studies have focused on training robots to identify optimal grasping locations during garment folding~\cite{xue2023unifolding, avigal2022speedfolding, iim}.

Cloth Funnels~\cite{canberk2023cloth} trained robots to align garments into predefined deformable configurations, simplifying the manipulation process for downstream folding. UniGarmentManip~\cite{wu2024unigarmentmanip} explored learning dense visual correspondences for garment manipulation with category-level generalization. FabricFlowNet~\cite{weng2022fabricflownet} developed a cloth manipulation policy that leverages flow as both input and action representation to enhance performance. Zhou et al.~\cite{zhou2021learning} and Lin et al.~\cite{lin2022learning} have proposed learning cloth dynamics to improve manipulation capabilities.

AdaFold~\cite{longhini2024adafold} employed a model-based feedback-loop framework to optimize folding trajectories in real-time. Our work takes an innovative approach by leveraging point cloud trajectories instead of key points for task planning. This method eliminates the need for prior category knowledge, enabling the development of a unified model capable of handling diverse garments seamlessly.

\subsection{Language-Conditioned Object Manipulation}

Language-conditioned manipulation has garnered significant attention due to its user-friendly interaction and strong generalization capabilities~\cite{mees2022matters, misra2016tell, hatori2018interactively, paxton2019prospection, tellex2011understanding}. Stengel et al.~\cite{stengel2022guiding} introduced a Transformer-based model for mapping natural language commands to robot actions in 3D tasks. CLIPort~\cite{shridhar2022cliport} combines CLIP's semantic understanding with Transporter's spatial precision for language-guided manipulation.

With recent advances in Large Language Models (LLMs), researchers have explored leveraging LLMs to integrate language understanding into cloth folding. GPT-Fabric~\cite{raval2024gpt} employs GPT to directly output actions, informing a robot where to grasp and pull a simple square-shaped fabric. Deng et al.~\cite{deng2024learning} introduced a method for language-conditioned manipulation that utilizes graph dynamics to improve the understanding and prediction of deformable object behaviors.

Our work addresses the problem of language-guided garment manipulation, which requires a deep understanding of diverse categories, shapes, and deformations.

\subsection{Foundation Model for Robotic Manipulation}

In recent years, foundation models have shown broad potential in robotics~\cite{hu2023toward, firoozi2023foundation, huang2024rekep, flip}. Models such as R3M~\cite{nair2022r3m}, SayCan~\cite{ahn2022can}, and the RT series~\cite{brohan2022rt, brohan2023rt, o2023open} have demonstrated strong generalization and few-shot transfer capabilities. However, these models primarily handle rigid and articulated objects and often struggle to manage deformable objects effectively due to their complex deformations.

Although models like GPT-Fabric~\cite{raval2024gpt} can handle certain deformable objects, they are less sensitive to intermediate states and rely heavily on manually predefined key points or vision-language models (VLMs) to interpret key points. The ManiFoundation model~\cite{xu2024manifoundation} formulates manipulation tasks as contact synthesis, allowing for effective action prediction across various objects, including deformable ones. This versatility makes the ManiFoundation model suitable as the action prediction component for manipulating different garments in our framework.

\begin{figure*}[t!]
\vspace{5pt}
  \begin{center}
   \includegraphics[width=1.0\linewidth]{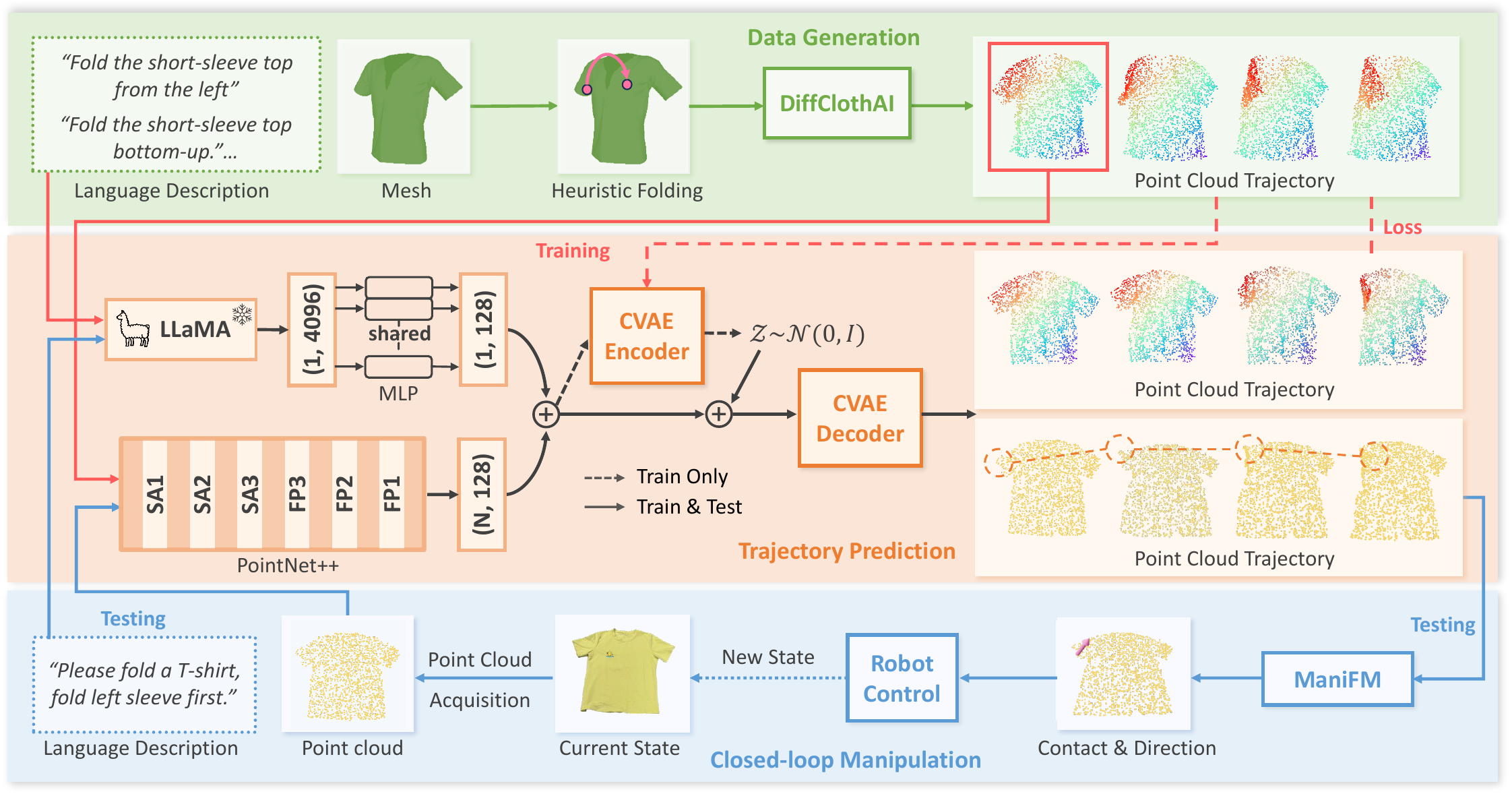}
  \end{center}
  \vspace{-5pt}
      \caption{Overview: The folding trajectory data for clothing is generated using heuristic methods in the DiffClothAI simulation environment, with language descriptions subsequently added \textbf{(Green)}. The trajectory generation model takes a point cloud from any given frame and a corresponding language description as inputs to generate the subsequent trajectory \textbf{(Orange)}. The generated trajectory is fed into the ManiFoundation model to estimate contact points and force directions, enabling the robot to conduct garment folding actions. This process is then iteratively refined using a feedback loop \textbf{(Blue)}.}
\label{fig:overview}
\end{figure*}

\section{Problem Formulation}
\label{sec:problem}

The goal of language-guided garment folding is to generate a sequence of actions $\{\boldsymbol{a}_i\}_{i=1}^n$ to fold the garment into a target point cloud configuration $\mathcal{P}_{goal}$, given the 3D point cloud observation $\mathcal{P}\in\mathbb{R}^{N\times3}$ with $N$ points and the language guidance $\mathcal{L}$.

As introduced before, we disentangle this task into three sub-tasks:
Point cloud trajectory generation, action prediction, and corresponding closed-loop manipulation:

(1) Given the current $\mathcal{P}$ with $N$ points and $\mathcal{L}$, the goal of the point cloud trajectory generation model is to generate the trajectory $\mathcal{T}=\{\mathcal{P}_i\}_{i=1}^{M}\in \mathbb{R}^{M\times N\times3}$ 
% \zhixuan{may better: MxNx3}\haonan{Done} 
that represents the evolution of the point cloud over time, where $M$ is the number of frames.

(2) Given two point clouds $(\mathcal{P}, \mathcal{P'})$, the aim of the ManiFoundation~\cite{xu2024manifoundation} model is to predict the action $\boldsymbol{a} = \{c_i\}_{i=1}^n$, which is defined as a set of contact syntheses. The contact synthesis for end effector $i$ is $c_i = (\boldsymbol{p}, \boldsymbol{s})$, where $\boldsymbol{p}\in\mathbb{R}^3$ is the contact position and $\boldsymbol{s}\in\mathbb{R}^3$ is the corresponding motion direction based on the trajectory.
% represented as point cloud P
% The garment folding task in our framework can be formulated as follows: (1) given a current garment observation $\mathcal{P}$ and user language guidance $\mathcal{L}$, the point cloud trajectory generation model generates a trajectory $\mathcal{T} = \mathcal{M}_{Gen}(\mathcal{P}, \mathcal{L})$. (2) 
We slice the generated garment point cloud trajectory $\mathcal{T}$ and input the segments into the ManiFoundation model to predict an action $\boldsymbol{a} = \mathcal{M}_{MF}(\mathcal{P}, \mathcal{P'})$, where $\{\mathcal{P}, \mathcal{P'}\} \subseteq \mathcal{T}$.

(3) Afterward, the robot executes the action $\boldsymbol{a}$, manipulating the garment to a new configuration $\mathcal{P}^* = \mathcal{M}_{Robot}(\mathcal{P}, \boldsymbol{a})$. 

We perform processes (1), (2), and (3) iteratively until the current point cloud configuration $\mathcal{P}$ matches the desired point cloud configuration $\mathcal{P}_{goal}$.

\section{Method}
\label{sec:method}

\subsection{Overview}

Our framework is illustrated in Figure~\ref{fig:overview}. In this section, we provide a detailed explanation of language-guided point cloud trajectory generation (Sec.~\ref{Method:Language-Guided Trajectory Generation}) and closed-loop manipulation (Sec.~\ref{Method:Closed-Loop Manipulation}).
Below, we first introduce the training and inference pipelines.

\textbf{Training.} 
During the model training process (green, orange), we generate a set of garment folding trajectory data for training (\ref{Method:Data Generation}). These data are used as ground truth 
to train the model, enabling it to produce outputs with a distribution similar to these data (\ref{Method:Conditional Point Cloud Trajectory Generation}). The generalization of language descriptions is discussed in~\ref{Method:Semantic Generalization}.

\textbf{Manipulation.} 
In the MetaFold Manipulation pipeline (orange, blue), we begin with Point Cloud Acquisition (\ref{Method:Point Cloud Acquisition}). The acquired point cloud is then fed into the trained model for inference, which outputs the point cloud trajectory. Segments of this trajectory are passed to the ManiFoundation model to predict the robot's actions (\ref{Method:ManiFoundation Model}). The robot executes these actions to fold the garment, followed by a new observation to continue the process (\ref{Method:FeedbackControl}).

\subsection{Language-Guided Trajectory Generation}
\label{Method:Language-Guided Trajectory Generation}

\subsubsection{Data Generation}
\label{Method:Data Generation}

We propose a framework that disentangles point cloud trajectory generation from action prediction, necessitating the training of a point cloud trajectory generation model. To the best of our knowledge, datasets that specifically provide point cloud trajectories for clothing folding are currently limited. We use ClothesNetM~\cite{Zhou_2023_ICCV} as the initial garment dataset, as it contains a wide range of well-formed garment models that support accurate simulation of deformation dynamics. Using DiffClothAI~\cite{10341573}, a differentiable simulator capable of handling deformable objects, we generate the point cloud trajectories for garment folding.

To define the grasping and target points, we employ a heuristic approach. We define a trajectory curve from the grasping point to the target point, with the simulator controlling the grasping point to move along this trajectory toward the target. The mesh vertices are extracted as point clouds and then downsampled to an appropriate size. This process generates ground-truth point cloud trajectories. Additionally, we add the corresponding language descriptions for different types of clothing and folding actions. 

\begin{figure}[t]
\vspace{5pt}
  \centering
  \includegraphics[width=1.0\linewidth]{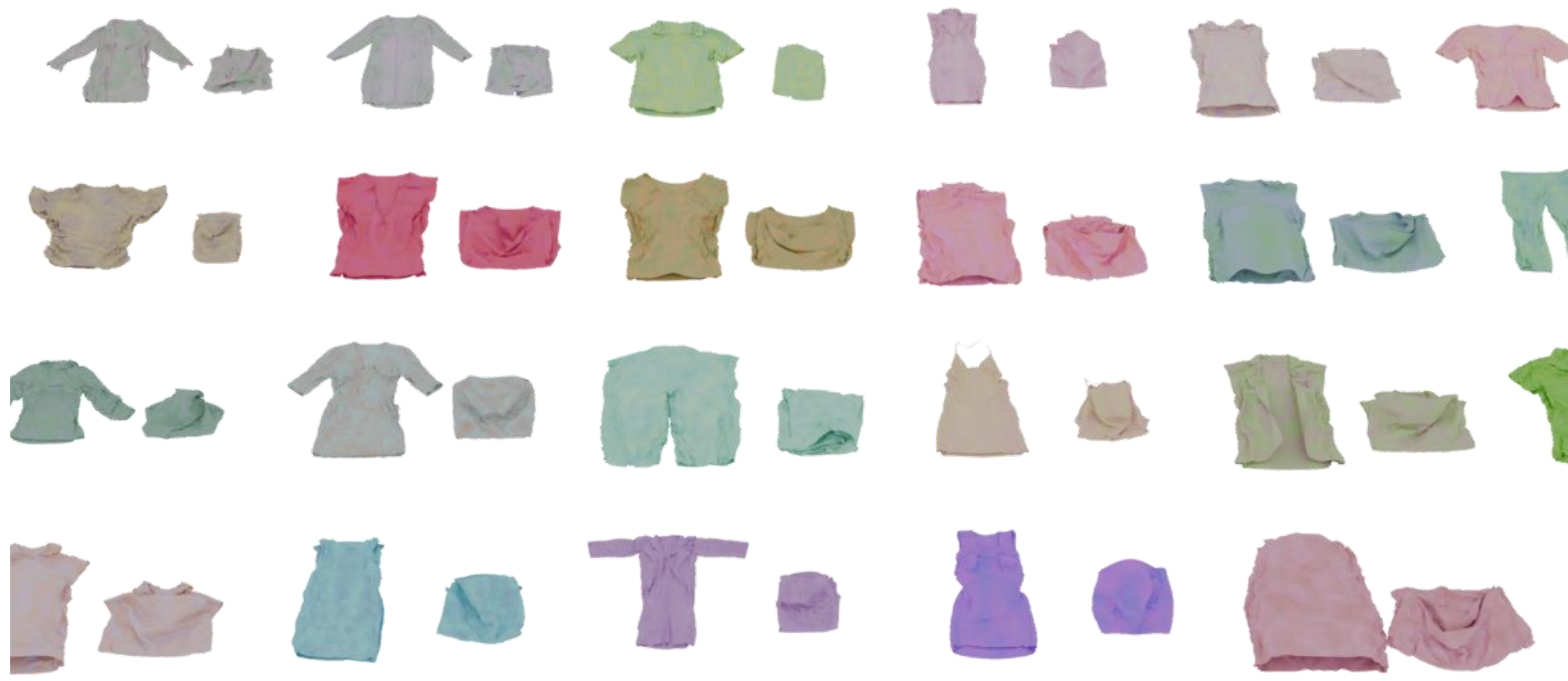}
  \vspace{-15pt}
  \caption{Example garments in our MetaFold Dataset. The garments demonstrate diversities in categories, shapes, and deformations.}
\label{fig:dataset}
\vspace{-10pt}
\end{figure}

Our dataset is visualized in Figure~\ref{fig:dataset}. The dataset consists of folding
point cloud trajectories from a total of 1210 garments and 3376 trajectories, with 2664 trajectories in the training set and 712 in the test set.

\subsubsection{Conditional Point Cloud Trajectory Generation}
\label{Method:Conditional Point Cloud Trajectory Generation}

As discussed earlier, point cloud trajectories can model state transitions, allowing us to control intermediate states during garment folding. To achieve this, we design a point cloud trajectory generation model as a world model. This approach enables a comprehensive generation and control of garment movement over time, providing a complete understanding of its dynamics. From a spatial perspective, this method tracks all points in the initial frame and maps the subsequent trajectory of each point. Temporally, each frame represents the garment's state at a specific moment.

Our point cloud trajectory generation model is built using a Conditional Variational Autoencoder (CVAE)~\cite{sohn2015learning}, where both the encoder and decoder are implemented using Transformer encoder architectures. Given that the point cloud $\mathcal{P}\in \mathbb{R}^{N\times 3}$ contains 
$N$ points and the language description $\mathcal{L}$, the goal of trajectory generation is to generate the point cloud trajectory $\mathcal{T}=\{\mathcal{P}_i\}_{i=1}^{M}\in \mathbb{R}^{N\times M\times3}$, where $M$ is the number of frames. PointNet++~\cite{qi2017pointnet++} is used to extract spatial information from the point cloud, resulting in a point cloud feature $\mathcal{F}_\mathcal{P} \in \mathbb{R}^{N\times 128}$. Meanwhile, LLaMA~\cite{touvron2023llamaopenefficientfoundation} processes the semantic information from the language description. We take the mean along the token direction to obtain the language feature with dimension $\mathbb{R}^{1\times 4096}$. To prevent the high-dimensional semantic features from dominating the spatial information, we use a multilayer perceptron (MLP) to reduce their dimensionality to 128, aligning them with the point cloud feature $\mathcal{F}_\mathcal{P}$. We denote the language feature as $\mathcal{F}_\mathcal{L} \in \mathbb{R}^{1\times 128}$.

\begin{figure}[!t]

    \centering
    \includegraphics[width=0.925\columnwidth]{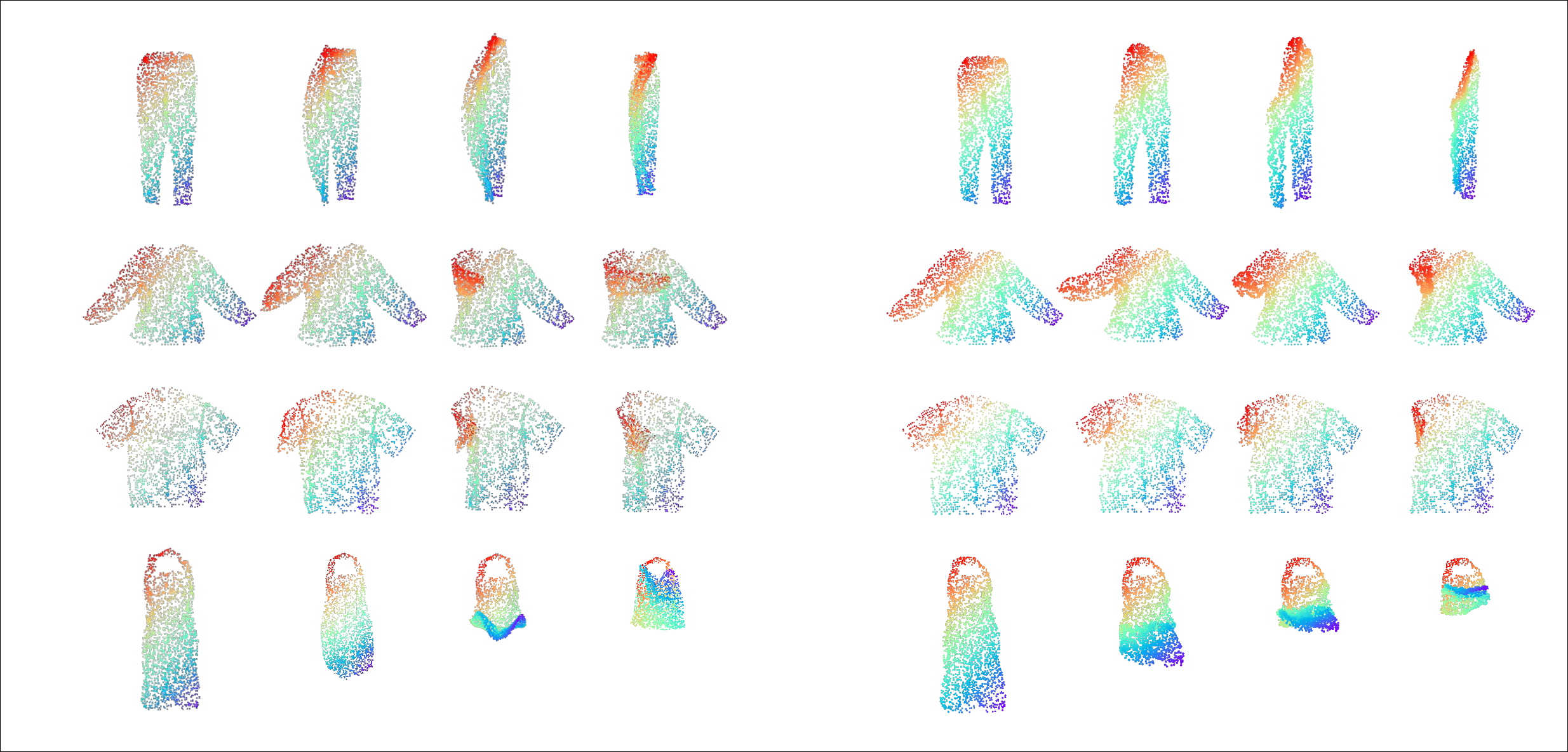}
    {\footnotesize \\ a) Ground truth folding trajectories \quad b) Generated folding trajectories}
    \vspace{-0.6em}
    \caption{Visualization of ground truth and generated trajectories.}
    \label{fig:Visualization of ground truth and generated trajectories.}
    \vspace{-7pt}
\end{figure}

In training, both the spatial features of the point cloud $\mathcal{F}_\mathcal{P}$ and the language features $\mathcal{F}_\mathcal{L}$ are combined with the ground truth trajectory and used as conditional inputs to the CVAE encoder. This optimizes the distribution of the latent variable $z$. In both training and inference, samples are drawn from the distribution of 
$z$ and combined with the spatial features of the point cloud $\mathcal{F}_\mathcal{P}$. The language features $\mathcal{F}_\mathcal{L}$ are used as conditional inputs to the CVAE decoder to generate trajectory features $\mathcal{F}_\mathcal{T} \in \mathbb{R}^{M\times N \times 256}$. An additional MLP is used to reduce the dimensionality of the output, providing the generated trajectory $\mathcal{T} \in \mathbb{R}^{M\times N\times 3}$ for the current frame. The visualization of generated trajectories is in Figure \ref{fig:Visualization of ground truth and generated trajectories.}.

It is worth noting that our model generates only one folding stage at a time (e.g., folding the left sleeve of a T-shirt), thereby focusing on a single subtask per iteration. This design advantageously allows for a clear separation of stages, simplifying the overall process while enabling the model to sequentially address different tasks by precisely guiding each folding stage. Consequently, this structure enhances generalization capabilities by accommodating adaptable, context-specific folding instructions across a variety of garment manipulation scenarios.

We train our model jointly rather than separately for different categories of garments. This approach allows our model to learn both the shared and distinct characteristics of various types of clothing. As a result, the model can automatically identify the appropriate folding method for each garment type and generate the corresponding folding trajectory, enabling cross-category adaptability.

\subsubsection{Semantic Generalization}
\label{Method:Semantic Generalization}

We use a pre-trained large language model (Meta-Llama-3.1-8B-Instruct)~\cite{touvron2023llamaopenefficientfoundation} to extract semantic features. Specifically, we predefine a set of garment folding descriptions to strengthen the encoding process. During inference, we also employ LLaMA as the foundational language model to process user-provided guidance. The model matches user input to our predefined descriptions, effectively reducing the complexity of training on diverse language instructions by constraining conditions within a defined range. This approach leverages the large language model to achieve open-vocabulary generalization.

\subsection{Closed-Loop Manipulation}
\label{Method:Closed-Loop Manipulation}

Our framework employs a closed-loop manipulation strategy that integrates point cloud acquisition, action prediction, and feedback control to achieve robust garment folding. This closed-loop design allows continuous adaptation to environmental variations and disturbances, ensuring reliable and precise performance during manipulation tasks.

\subsubsection{Point Cloud Acquisition}
\label{Method:Point Cloud Acquisition}

In the simulation, we can conveniently obtain current point clouds by extracting the mesh vertices of the garment. In the real world, we use an RGB-D camera to capture both RGB images and depth maps. The clothing point cloud can be obtained through the Segment Anything Model 2 (SAM 2)~\cite{ravi2024sam}. This point cloud is then downsampled to a dimension suitable for the trajectory generation model and the ManiFoundation model~\cite{xu2024manifoundation}. Additionally, users can specify garment folding methods via different language descriptions, such as indicating whether to fold the left sleeve or the right sleeve first.

\subsubsection{ManiFoundation Model}
\label{Method:ManiFoundation Model}

Using the trajectory generation model, we generate point cloud trajectories, obtaining the flow between successive point clouds. The ManiFoundation model takes this flow between two point clouds as input and predicts a series of contact syntheses, which represent the actions the robot needs to perform. By disentangling task understanding and action prediction, we only need to train the trajectory generation model and the ManiFoundation model separately. This modular design simplifies training and enhances adaptability for generalization.

We fine-tuned the ManiFoundation model to enhance its performance in action prediction for garment folding. We leveraged a subset of garment data and conducted folding tasks. During fine-tuning, we applied a loss function that measured the discrepancy between the model's predicted contact syntheses and ground truth. This process allowed us to optimize the model’s ability to accurately predict contact points and force directions, thereby improving its overall performance in garment folding tasks.

Additionally, the predictions from the ManiFoundation model are influenced by random seeds. To mitigate this variability, we employ a model ensemble approach using 160 different random seeds to generate multiple predictions. When the distance between two prediction results is less than $\varepsilon$, they are considered similar and grouped. If a prediction result does not match any existing group, a new group is created. We compute the arithmetic mean of the point positions within the modal group and select the point closest to this average, along with its corresponding force, as the output contact point and force direction.

\subsubsection{Feedback Control}
\label{Method:FeedbackControl}

After receiving the action output from the ManiFoundation model, the robot's end effector grasps and moves the garment according to the specified action. After a short interval, the garment reaches a new state, and we capture its current point cloud. 
The point cloud is then fed into the trajectory generation model to produce the subsequent trajectory. This process establishes a closed-loop feedback control, allowing our framework to adapt to disturbances and variations in the environment.

\section{Experiments}
\label{sec:experiments}

\begin{table*}[tbp]
\vspace{5pt}
  \caption{Simulation results on garment folding tasks. UniG stands for UniGarmentManip~\cite{wu2024unigarmentmanip}.}
  \vspace{-5pt}
  \label{tab:Folding Results}
  \centering
  
  \begin{adjustbox}{max width=\textwidth}
  \begin{tabular}{lcccccccccc} % cccccccccc llllllllll
    \toprule
    {} & {} & \multicolumn{4}{c}{{MetaFold Dataset}} & \multicolumn{4}{c}{{Cloth3D}} \\
    \cmidrule(lr){3-6} \cmidrule(lr){7-10}
    {} & {} & {No-sleeve} & {Short-sleeve} & {Long-sleeve} & {Pants} & {No-sleeve} & {Short-sleeve} & {Long-sleeve} & {Pants} \\
    \midrule

    % \multirow{4}{*}{IoU $\uparrow$} & \UniG   & 0.72 & 0.57 & \textbf{0.72} & 0.57 & - & - & - & - \\
    %                     & \UniG (multi)   & 0.19 & 0.18 & 0.46 & 0.45 & - & - & - & - \\
    %                     & GPT-Fabric   & 0.57 & 0.38 & 0.48 & 0.59 & - & - & - & - \\
    %                      & \Ours   & \textbf{0.79} & \textbf{0.58} & 0.55 & \textbf{0.65} & - & - & - & - \\ \midrule
    \multirow{4}{*}{Rectangularity $\uparrow$} & UniG   &0.85 &0.78 &\textbf{0.88} &0.81 &\textbf{0.82} &\textbf{0.80 }&\textbf{0.85 }&0.83 \\
                         % & UniG (multi)   &0.85 &0.82 &0.87 &0.85 &\textbf{0.83} &\textbf{0.83} &\textbf{0.86} &\textbf{0.83} \\
                         & DP3   &0.85 &0.82 &0.86 &\textbf{0.88} &0.80 &0.76 &0.78 &0.79 \\
                         & GPT-Fabric   & 0.78 & 0.78 & 0.77 & 0.66 & 0.81 & 0.78 & 0.80 & \textbf{0.83} \\
                         & Ours   & \textbf{0.87} & \textbf{0.83} & 0.85 & 0.86 & \textbf{0.82} & \textbf{0.80} & 0.83 & \textbf{0.83} \\ \midrule
    \multirow{4}{*}{Area Ratio $\downarrow$} & UniG   &0.48 &0.34 &0.34 &0.34 &\textbf{0.47} &0.43 &0.34 &0.28 \\
                         % & UniG (multi)  &0.48 &\textbf{0.30} &0.32 &0.33 &\textbf{0.47} &0.37 &0.32 &0.29 \\
                         & DP3  &0.50 &0.44 &0.39 &0.33 &\textbf{0.47} &\textbf{0.33} &0.26 &0.28 \\
                         & GPT-Fabric   & 0.48 & 0.45 & 0.47 & 0.44 & 0.54 & 0.46 & 0.47 & 0.50 \\
                         & Ours   & \textbf{0.45} & \textbf{0.33} & \textbf{0.24} & \textbf{0.26} & \textbf{0.47} & \textbf{0.33} & \textbf{0.25} & \textbf{0.27} \\ \midrule
    \multirow{4}{*}{Success Rate $\uparrow$} & UniG   &0.71 &0.69 &\textbf{0.90} &0.77 &0.77 &0.42 &0.71 &0.91 \\
                         % & UniG (multi) &0.71 &0.72 &\textbf{0.90} &0.86 &\textbf{0.79} &0.62 &0.80 &0.90 \\
                         & DP3  &0.73 &0.66 &0.37 &0.94 &0.71 &0.70 &0.82 &0.85 \\
                         & GPT-Fabric   & 0.34 & 0.21 & 0.03 & 0.40 & 0.63 & 0.22 & 0.15 & 0.03 \\
                         & Ours   & \textbf{0.88} & \textbf{0.86} & \textbf{0.90} & \textbf{0.97} & \textbf{0.79} & \textbf{0.86} & \textbf{0.97} & \textbf{0.97} \\
    
    \bottomrule
  \end{tabular}
  \end{adjustbox}
\vspace{-8pt}
\end{table*}

In this section, we aim to address the following questions:

\begin{enumerate}[label=Q\arabic*:, leftmargin=2em, labelwidth=1.5em, itemindent=!, align=left]
    \item How does our model's performance on garment folding tasks compare with baselines?
    \item How much better does our approach of disentangling task planning and actions perform than the end-to-end method that directly predicts actions?
    \item How well does our model generalize to different language descriptions and folding methods?
\end{enumerate}

\begin{table}[!tbp]
  % \vspace{-5pt}
  % \caption{Language Experiments. \Language is the baseline in our paper.}
  \caption{Different language-guided folding tasks. L.D. stands for ~\cite{deng2024learning}. "Seen" and "Unseen" refer to the input instructions.}
  \vspace{-5pt}
  \label{tab:Language Guidance results}
  \centering
  \small
  % \vspace{-10pt}
  % {\def\arraystretch{0.6}       % 缩短表格内部上下间距
  \begin{tabular}{lccccc} % cccccccccc llllllllll
    \toprule
    {} & {} & \multicolumn{2}{c}{{MetaFold Dataset}} & \multicolumn{2}{c}{{Cloth3D}} \\
    \cmidrule(lr){3-4} \cmidrule(lr){5-6}
    {} & {} & {Seen} & {Unseen}  & {Seen} & {Unseen}  \\
    \midrule

    % \multirow{2}{*}{IoU $\uparrow$} &\Language   & \textbf{0.62 } & 0.60  & - & - \\
    %                      & \textbf{\Ours}   & 0.55  & \textbf{0.63 }  & - & -  \\ \midrule
    \multirow{2}{*}{Rectangularity $\uparrow$} &L.D.  & 0.78  & 0.78 & 0.81  & \textbf{0.81}  \\
                         & \textbf{Ours}   & \textbf{0.85} & \textbf{0.80} & \textbf{0.83} & \textbf{0.81} \\ \midrule
    \multirow{2}{*}{Area Ratio $\downarrow$} &L.D.   & 0.36 & 0.37 & 0.39& 0.40 \\
                         & \textbf{Ours}   & \textbf{0.24} & \textbf{0.33} & \textbf{0.25} & \textbf{0.26} \\ \midrule
    \multirow{2}{*}{Success Rate $\uparrow$} &L.D.   & 0.46 & 0.46 & 0.56 & 0.47 \\
                         & \textbf{Ours}   & \textbf{0.90} & \textbf{0.63} & \textbf{0.97} & \textbf{0.93} \\
    
    \bottomrule
  \end{tabular}
  % }
\vspace{-10pt}
\end{table}

\subsection{Simulation and Dataset}

We use Isaac Sim~\cite{liang2018gpu} as the simulation environment of our framework. Compared to PyFleX~\cite{li2018learning, wu2024unigarmentmanip} used by baselines, it provides a more accurate simulation of internal forces in garments, resulting in higher accuracy for fabric simulation and lower occurrence of mesh penetration.

We conducted experiments using two datasets. The first is our own generated trajectory dataset (Sec.~\ref{Method:Data Generation}), called the \textbf{MetaFold dataset}, from which we take the test set for evaluation. The garment meshes in this dataset are provided by ClothesNet~\cite{Zhou_2023_ICCV}. The second dataset is CLOTH3D~\cite{bertiche2020cloth3d}, on which our model is evaluated in a \textbf{zero-shot} setting. We tested approximately 500 different garments in total.

\subsection{Metrics}

We employed multiple metrics to evaluate the folding quality from various perspectives.

\textbf{Rectangularity.} The ratio of the final garment area to its bounding rectangle. We use this metric to assess how well the folded garment resembles a rectangle, providing a measure of the folding quality.

\textbf{Area Ratio.} The ratio of the final garment area to the initial garment area. This metric indicates the folding quality, as each fold that brings points closer to garment boundaries reduces the area, leading to a more compact configuration.

\textbf{Success Rate.} The success rate is measured by the rectangularity exceeding a specified threshold and the area ratio falling below a certain threshold.

\subsection{Baselines}

For the garment folding task (Q1), we use UniGarmentManip~\cite{wu2024unigarmentmanip} as the baseline. This method, previously demonstrated state-of-the-art (SOTA) performance in individual garment folding tasks~\cite{wu2024unigarmentmanip}, supports folding various garment types, thereby enabling a comprehensive comparison of folding performance across different categories. We use GPT-Fabric~\cite{raval2024gpt} as a baseline, which is a recent method that leverages large language models to select key points.

For end-to-end method comparison (Q2), we use the 3D Diffusion Policy (DP3)~\cite{Ze2024DP3} as a baseline, which directly outputs the next action from the input point cloud.

For the language guidance task (Q3), we use~\cite{deng2024learning} as a baseline. This baseline, previously the SOTA in language-guided garment folding~\cite{deng2024learning}, supports diverse language inputs to guide garment folding. We evaluate the generalization ability of the models by comparing their performance across different language descriptions, demonstrating our approach's robust language comprehension and adaptability.

\subsection{Results and Analysis}

Table~\ref{tab:Folding Results} presents the results for garment folding tasks, Table~\ref{tab:Language Guidance results} shows the results for different language guidance, and Table~\ref{tab:ablation_results2} provides ablation results. In the tables, $\uparrow$ indicates that higher values are better, while $\downarrow$ indicates that lower values are preferable.

\textbf{Garment Folding.} In the garment folding task, UniGarmentManip demonstrates reasonable performance by using separate checkpoints and demonstrations for different garments. However, its heavy reliance on demonstrations can sometimes lead to geometric failures, resulting in poor folding outcomes for some garments. GPT-Fabric, which relies on VLMs and LLMs for keypoint selection, often struggles to select accurate key points, making it less effective for complex tasks like garment folding. Table~\ref{tab:Folding Results} presents the quantitative results of our method compared to these baselines on different garment folding tasks.

Although our approach is zero-shot on the CLOTH3D dataset, we achieve performance that matches or even surpasses the baseline, demonstrating the strong generalization of our method to a wide range of new garments. In garment folding, our model tends to fold clothing into a small and compact form, which aligns well with human expectations for neatness and compactness. In general, our approach outperforms the baselines in most tasks and metrics. Therefore, we can address Q1, demonstrating our superior performance in garment folding.

In addition, our approach, which disentangles task planning from action prediction, achieves better results than the end-to-end method DP3~\cite{Ze2024DP3}, which directly predicts actions. By separating high-level planning from low-level control, our method can better capture the underlying complexities of garment manipulation, leading to more precise action sequences. This modular strategy enhances learning efficiency and robustness, ultimately addressing Q2.

\begin{figure}[t]
    \centering
    \vspace{-3pt}
    \includegraphics[width=0.9\linewidth]{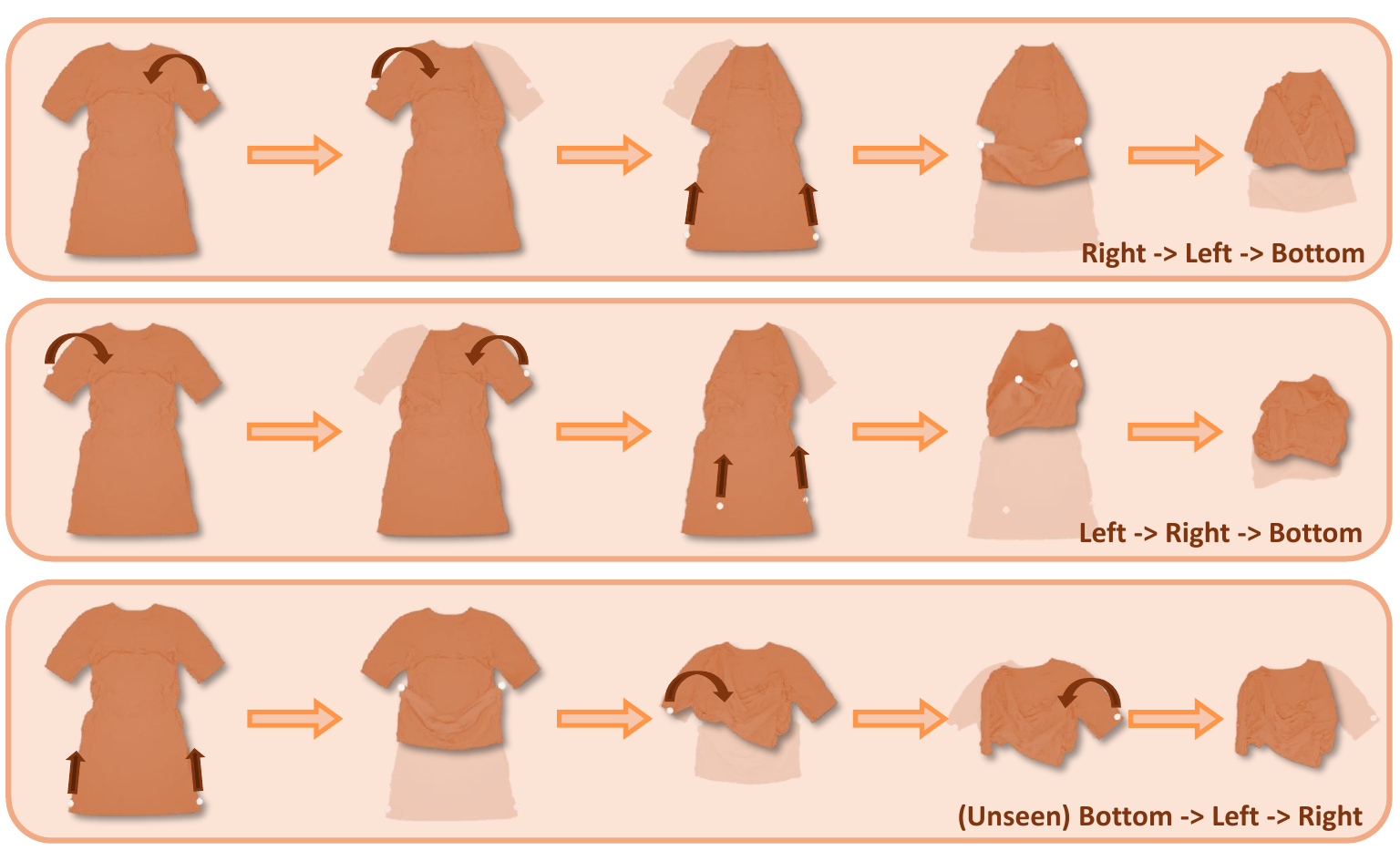}
    \vspace{-5pt}
    \caption{Guided by different languages, our framework can generate different corresponding garment folding sequences.}
    \label{fig:sim_visPS}
    \vspace{-5pt}
\end{figure}

\begin{figure*}[t!]
    \centering
    \includegraphics[width=0.8\textwidth]{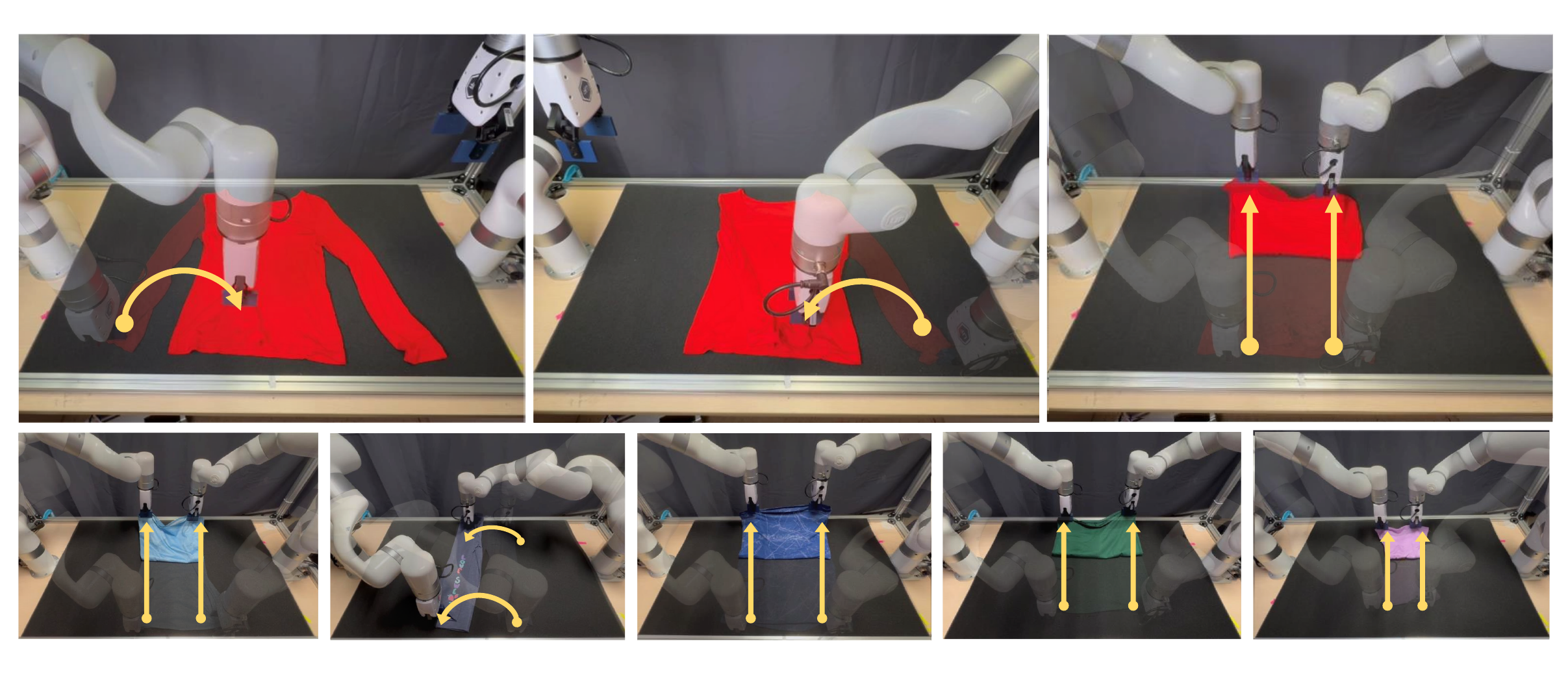}
    \vspace{-20pt}
    \caption{Real-World Experiments of MetaFold on Diverse Garment Types: Long-Sleeve, Short-Sleeve, No-Sleeve, and Pants.}
    \label{fig:real_world}
\end{figure*}

 \begin{table*}[tbp!]
    \caption{Ablation Studies. All ablation experiments were conducted on the same garment type.}
    \vspace{-5pt}
    \label{tab:ablation_results2}
    \centering
    \begin{adjustbox}{max width=\textwidth}
    \begin{tabular}{lcccccc}
      \toprule
      Methods & \textbf{Ours}  & Ours-5frames & Ours-15frames & Ours-NextStep  & Ours w/o MF & Ours w/o CL \\
      \midrule
       % IoU $\uparrow$ & 0.58 & 0.54 & \textbf{0.61} & 0.49\\
       Rectangularity $\uparrow$ & \textbf{0.83} & 0.80 & \textbf{0.83} & 0.79& 0.81 & 0.81\\
       Area Ratio $\downarrow$ & \textbf{0.33} & 0.42 & 0.34 & 0.35& 0.46 & 0.60 \\
       Success Rate $\uparrow$ & \textbf{0.86 }& 0.51 & 0.69 & 0.41& 0.27 & 0.07 \\
      \bottomrule
    \end{tabular}
    \end{adjustbox}
\vspace{-10pt}    
\end{table*}%

\textbf{Language-Guided Garment Folding.} In the language-guided garment folding task, we compared different levels of language generalization. Seen instructions refer to the guidance that our model and~\cite{deng2024learning} encountered during their respective training processes. Unseen instructions are guidance phrases that neither model has encountered before, yet they direct the same folding tasks. Table~\ref{tab:Language Guidance results} presents the results of our method compared to~\cite{deng2024learning} across these language scenarios. Under both instruction types, our model demonstrates superior folding performance compared to the baseline. This addresses Q3, demonstrating that our model has strong language understanding and generalization capabilities for garment folding tasks.

Furthermore, our model supports folding based on user-provided language inputs, even if the folding sequence (Bottom $\rightarrow$ Left $\rightarrow$ Right) has not appeared in the training data. Figure~\ref{fig:sim_visPS} illustrates the folding results of our model under different language instructions.

\subsection{Ablation Studies}

We compared our framework with several ablated versions to demonstrate the effectiveness of its components:
\begin{itemize}
    \item \textbf{Ours w/o MF}: our method without ManiFoundation model. We randomly select points from the entire set, filter out those with minimal motion, and use grouping to determine contact points. We use the selected point's trajectory direction prediction as the force direction. 
    
    \item \textbf{Ours w/o CL}: our method uses open-loop control instead of closed-loop control. The execution relies entirely on the initial frame's predicted trajectory without re-predicting in later steps.
    
    \item \textbf{Ours-5frames and Ours-15frames}: represent different granularities of closed-loop execution, performed every 5 frames and every 15 frames, respectively.

    \item \textbf{Ours-NextStep}: predicts only a single step at a time rather than an entire trajectory. 
\end{itemize}

Table~\ref{tab:ablation_results2} presents a quantitative comparison with these ablated versions. Experimental results indicate that both the ManiFoundation model and closed-loop control are essential components in garment folding tasks. The results also indicate that our approach achieves optimal performance when closed-loop execution is performed every 10 frames. Predicting the entire trajectory enables the model to generate more effective sequences of actions.

\subsection{Real-World Experiments}

We conducted real-robot experiments using a uFactory xArm6 robot equipped with a uFactory xArm Gripper and an overhead RealSense D435 camera. The settings are the same as in Fig.~\ref{fig:teaser}.

To obtain the garment point cloud, we first segmented the RGB image captured by an RGB-D camera using SAM2~\cite{ravi2024sam2} to generate a garment mask. This mask was then used to filter the corresponding depth data, extracting the actual point cloud of the clothing. The resulting points are then fed directly into our framework to forecast future folding states and the directions of contact-point movements. By executing these actions in sequence, the garment-folding process can be carried out in the real world. Because the sim-to-real gap in the point clouds is small, our approach can be transferred directly to real-world settings. The qualitative results are shown in Fig.~\ref{fig:real_world}, and the quantitative outcomes are listed in Table~\ref{tab:Real World Results}, with each garment tested over 10 trials.

\begin{table}[!tbp]
  % \vspace{-5pt}
  % \caption{Language Experiments. \Language is the baseline in our paper.}
  \caption{Real world results for our framework}
  \vspace{-5pt}
  \label{tab:Real World Results}
  \centering
  \small
  % \vspace{-10pt}
  % {\def\arraystretch{0.6}       % 缩短表格内部上下间距
  \begin{adjustbox}{max width=\linewidth}
  \begin{tabular}{lccccc} % cccccccccc llllllllll
    \toprule
    {}  & {No-sleeve} & {Short-sleeve}  & {Long-sleeve} & {Pants}  \\
    \midrule

    % \multirow{2}{*}{IoU $\uparrow$} &\Language   & \textbf{0.62 } & 0.60  & - & - \\
    %                      & \textbf{\Ours}   & 0.55  & \textbf{0.63 }  & - & -  \\ \midrule
    \multirow{1}{*}{Rectangularity $\uparrow$}  & 0.94  & 0.91 & 0.87  & 0.85 \\ 
    \multirow{1}{*}{Area Ratio $\downarrow$}  & 0.45  & 0.33 & 0.29  & 0.24 \\ 
    \multirow{1}{*}{Success Rate $\uparrow$} & 10/10  & 8/10 & 9/10  & 9/10 \\
    
    \bottomrule
  \end{tabular}
  \end{adjustbox}
  % }
\vspace{-10pt}
\end{table}
\section{Conclusion}
\label{sec:Conclusion}

In this work, we propose a comprehensive framework, \textbf{MetaFold}, for garment folding that supports multi-category folding tasks guided by user language instructions. This framework adopts a disentangled structure consisting of a point cloud trajectory generation model and a low-level action prediction model, utilizing closed-loop control for effective garment manipulation. We also constructed a point cloud trajectory dataset for garment folding, encompassing various folding methods across different garment types. Experimental results demonstrate that our approach achieves state-of-the-art performance in multi-category and language-guided garment folding tasks. We believe that MetaFold represents a significant step forward in applying trajectory generation to deformable object manipulation, marking an important milestone toward advancing spatial intelligence.
\section*{Acknowledgements}

This research / project is supported by the Ministry of Education, Singapore, under the Academic Research Fund Tier 1 (FY2024).

% \addtolength{\textheight}{-12cm} 
\bibliographystyle{IEEEtran}
\bibliography{IEEEabrv,ref}

\begin{thebibliography}{10}
\providecommand{\url}[1]{#1}
\csname url@rmstyle\endcsname
\providecommand{\newblock}{\relax}
\providecommand{\bibinfo}[2]{#2}
\providecommand\BIBentrySTDinterwordspacing{\spaceskip=0pt\relax}
\providecommand\BIBentryALTinterwordstretchfactor{4}
\providecommand\BIBentryALTinterwordspacing{\spaceskip=\fontdimen2\font plus
\BIBentryALTinterwordstretchfactor\fontdimen3\font minus \fontdimen4\font\relax}
\providecommand\BIBforeignlanguage[2]{{%
\expandafter\ifx\csname l@#1\endcsname\relax
\typeout{** WARNING: IEEEtran.bst: No hyphenation pattern has been}%
\typeout{** loaded for the language `#1'. Using the pattern for}%
\typeout{** the default language instead.}%
\else
\language=\csname l@#1\endcsname
\fi
#2}}

\bibitem{zhu2022challenges}
J.~Zhu, A.~Cherubini, C.~Dune, D.~Navarro-Alarcon, F.~Alambeigi, D.~Berenson, \emph{et~al.}, ``Challenges and outlook in robotic manipulation of deformable objects,'' \emph{IEEE Robotics \& Automation Magazine}, 2022.

\bibitem{gu2023survey}
F.~Gu, Y.~Zhou, Z.~Wang, S.~Jiang, and B.~He, ``A survey on robotic manipulation of deformable objects: Recent advances, open challenges and new frontiers,'' \emph{arXiv preprint arXiv:2312.10419}, 2023.

\bibitem{shi2021skeleton}
R.~Shi, Z.~Xue, Y.~You, and C.~Lu, ``Skeleton merger: an unsupervised aligned keypoint detector,'' in \emph{CVPR}, 2021, pp. 43--52.

\bibitem{ganapathi2021learning}
A.~Ganapathi, P.~Sundaresan, B.~Thananjeyan, A.~Balakrishna, D.~Seita, J.~Grannen, \emph{et~al.}, ``Learning dense visual correspondences in simulation to smooth and fold real fabrics,'' in \emph{ICRA}, 2021.

\bibitem{wu2024unigarmentmanip}
R.~Wu, H.~Lu, Y.~Wang, Y.~Wang, and H.~Dong, ``Unigarmentmanip: A unified framework for category-level garment manipulation via dense visual correspondence,'' in \emph{CVPR}, 2024, pp. 16\,340--16\,350.

\bibitem{hietala2022learning}
J.~Hietala, D.~Blanco-Mulero, G.~Alcan, and V.~Kyrki, ``Learning visual feedback control for dynamic cloth folding,'' in \emph{IROS}.\hskip 1em plus 0.5em minus 0.4em\relax IEEE, 2022.

\bibitem{he2024fabricfolding}
C.~He, L.~Meng, Z.~Sun, J.~Wang, and M.~Q.-H. Meng, ``Fabricfolding: learning efficient fabric folding without expert demonstrations,'' \emph{Robotica}, vol.~42, no.~4, pp. 1281--1296, 2024.

\bibitem{canberk2023cloth}
A.~Canberk, C.~Chi, H.~Ha, B.~Burchfiel, E.~Cousineau, S.~Feng, and S.~Song, ``Cloth funnels: Canonicalized-alignment for multi-purpose garment manipulation,'' in \emph{ICRA}.\hskip 1em plus 0.5em minus 0.4em\relax IEEE, 2023, pp. 5872--5879.

\bibitem{huang2024rekep}
W.~Huang, C.~Wang, Y.~Li, R.~Zhang, and L.~Fei-Fei, ``Rekep: Spatio-temporal reasoning of relational keypoint constraints for robotic manipulation,'' \emph{arXiv preprint arXiv:2409.01652}, 2024.

\bibitem{raval2024gpt}
V.~Raval, E.~Zhao, H.~Zhang, S.~Nikolaidis, and D.~Seita, ``Gpt-fabric: Folding and smoothing fabric by leveraging pre-trained foundation models,'' \emph{arXiv preprint arXiv:2406.09640}, 2024.

\bibitem{xu2024manifoundation}
Z.~Xu, C.~Gao, Z.~Liu, G.~Yang, C.~Tie, H.~Zheng, H.~Zhou, \emph{et~al.}, ``Manifoundation model for general-purpose robotic manipulation of contact synthesis with arbitrary objects and robots,'' in \emph{IROS}, 2024.

\bibitem{hoque2022learning}
R.~Hoque, K.~Shivakumar, S.~Aeron, G.~Deza, A.~Ganapathi, A.~Wong, J.~Lee, A.~Zeng, V.~Vanhoucke, and K.~Goldberg, ``Learning to fold real garments with one arm: A case study in cloud-based robotics research,'' in \emph{IROS}, 2022.

\bibitem{lee2021learning}
R.~Lee, D.~Ward, V.~Dasagi, A.~Cosgun, J.~Leitner, and P.~Corke, ``Learning arbitrary-goal fabric folding with one hour of real robot experience,'' in \emph{CoRL}.\hskip 1em plus 0.5em minus 0.4em\relax PMLR, 2021, pp. 2317--2327.

\bibitem{tanaka2018emd}
D.~Tanaka, S.~Arnold, and K.~Yamazaki, ``Emd net: An encode--manipulate--decode network for cloth manipulation,'' \emph{RAL}, 2018.

\bibitem{stria2014garment}
J.~Stria, D.~Pr{\v{u}}{\v{s}}a, V.~Hlav{\'a}{\v{c}}, L.~Wagner, V.~Petrik, P.~Krsek, and V.~Smutn{\`y}, ``Garment perception and its folding using a dual-arm robot,'' in \emph{IROS}.\hskip 1em plus 0.5em minus 0.4em\relax IEEE, 2014, pp. 61--67.

\bibitem{xue2023unifolding}
H.~Xue, Y.~Li, W.~Xu, H.~Li, D.~Zheng, and C.~Lu, ``Unifolding: Towards sample-efficient, scalable, and generalizable robotic garment folding,'' \emph{arXiv preprint arXiv:2311.01267}, 2023.

\bibitem{avigal2022speedfolding}
Y.~Avigal, L.~Berscheid, T.~Asfour, T.~Kr{\"o}ger, and K.~Goldberg, ``Speedfolding: Learning efficient bimanual folding of garments,'' in \emph{IROS}.\hskip 1em plus 0.5em minus 0.4em\relax IEEE, 2022, pp. 1--8.

\bibitem{iim}
C.~Gao, Z.~Li, H.~Gao, and F.~Chen, ``Iterative interactive modeling for knotting plastic bags,'' in \emph{CoRL}.\hskip 1em plus 0.5em minus 0.4em\relax PMLR, 2023, pp. 571--582.

\bibitem{weng2022fabricflownet}
T.~Weng, S.~M. Bajracharya, Y.~Wang, K.~Agrawal, and D.~Held, ``Fabricflownet: Bimanual cloth manipulation with a flow-based policy,'' in \emph{CoRL}.\hskip 1em plus 0.5em minus 0.4em\relax PMLR, 2022, pp. 192--202.

\bibitem{zhou2021learning}
P.~Zhou, O.~Zahra, A.~Duan, S.~Huo, Z.~Wu, and D.~Navarro-Alarcon, ``Learning cloth folding tasks with refined flow based spatio-temporal graphs,'' \emph{arXiv preprint arXiv:2110.08620}, 2021.

\bibitem{lin2022learning}
X.~Lin, Y.~Wang, Z.~Huang, and D.~Held, ``Learning visible connectivity dynamics for cloth smoothing,'' in \emph{CoRL}, 2022, pp. 256--266.

\bibitem{longhini2024adafold}
A.~Longhini, M.~C. Welle, Z.~Erickson, and D.~Kragic, ``Adafold: Adapting folding trajectories of cloths via feedback-loop manipulation,'' \emph{arXiv preprint arXiv:2403.06210}, 2024.

\bibitem{mees2022matters}
O.~Mees, L.~Hermann, and W.~Burgard, ``What matters in language conditioned robotic imitation learning over unstructured data,'' \emph{RAL}, vol.~7, no.~4, pp. 11\,205--11\,212, 2022.

\bibitem{misra2016tell}
D.~K. Misra, J.~Sung, K.~Lee, and A.~Saxena, ``Tell me dave: Context-sensitive grounding of natural language to manipulation instructions,'' \emph{IJRR}, vol.~35, no. 1-3, pp. 281--300, 2016.

\bibitem{hatori2018interactively}
J.~Hatori, Y.~Kikuchi, S.~Kobayashi, K.~Takahashi, Y.~Tsuboi, Y.~Unno, W.~Ko, and J.~Tan, ``Interactively picking real-world objects with unconstrained spoken language instructions,'' in \emph{ICRA}.\hskip 1em plus 0.5em minus 0.4em\relax IEEE, 2018.

\bibitem{paxton2019prospection}
C.~Paxton, Y.~Bisk, J.~Thomason, A.~Byravan, and D.~Foxl, ``Prospection: Interpretable plans from language by predicting the future,'' in \emph{ICRA}.\hskip 1em plus 0.5em minus 0.4em\relax IEEE, 2019, pp. 6942--6948.

\bibitem{tellex2011understanding}
S.~Tellex, T.~Kollar, S.~Dickerson, M.~Walter, A.~Banerjee, S.~Teller, and N.~Roy, ``Understanding natural language commands for robotic navigation and mobile manipulation,'' in \emph{AAAI}, 2011.

\bibitem{stengel2022guiding}
E.~Stengel-Eskin, A.~Hundt, Z.~He, A.~Murali, N.~Gopalan, M.~Gombolay, and G.~Hager, ``Guiding multi-step rearrangement tasks with natural language instructions,'' in \emph{CoRL}, 2022, pp. 1486--1501.

\bibitem{shridhar2022cliport}
M.~Shridhar, L.~Manuelli, and D.~Fox, ``Cliport: What and where pathways for robotic manipulation,'' in \emph{CoRL}, 2022, pp. 894--906.

\bibitem{deng2024learning}
Y.~Deng, K.~Mo, C.~Xia, and X.~Wang, ``Learning language-conditioned deformable object manipulation with graph dynamics,'' in \emph{ICRA}.\hskip 1em plus 0.5em minus 0.4em\relax IEEE, 2024, pp. 7508--7514.

\bibitem{hu2023toward}
Y.~Hu, Q.~Xie, V.~Jain, J.~Francis, J.~Patrikar, N.~Keetha, S.~Kim, \emph{et~al.}, ``Toward general-purpose robots via foundation models: A survey and meta-analysis,'' \emph{arXiv preprint arXiv:2312.08782}, 2023.

\bibitem{firoozi2023foundation}
R.~Firoozi, J.~Tucker, S.~Tian, A.~Majumdar, J.~Sun, W.~Liu, Y.~Zhu, S.~Song, A.~Kapoor, K.~Hausman, \emph{et~al.}, ``Foundation models in robotics: Applications, challenges, and the future,'' \emph{IJRR}, 2023.

\bibitem{flip}
C.~Gao, H.~Zhang, Z.~Xu, C.~Zhehao, and L.~Shao, ``Flip: Flow-centric generative planning for general-purpose manipulation tasks,'' in \emph{CoRL Workshop on Learning Effective Abstractions for Planning}, 2024.

\bibitem{nair2022r3m}
S.~Nair, A.~Rajeswaran, V.~Kumar, C.~Finn, and A.~Gupta, ``R3m: A universal visual representation for robot manipulation,'' \emph{CoRL}, 2022.

\bibitem{ahn2022can}
M.~Ahn, A.~Brohan, N.~Brown, Y.~Chebotar, O.~Cortes, B.~David, C.~Finn, \emph{et~al.}, ``Do as i can, not as i say: Grounding language in robotic affordances,'' \emph{arXiv preprint arXiv:2204.01691}, 2022.

\bibitem{brohan2022rt}
A.~Brohan, N.~Brown, J.~Carbajal, Y.~Chebotar, J.~Dabis, C.~Finn, K.~Gopalakrishnan, \emph{et~al.}, ``Rt-1: Robotics transformer for real-world control at scale,'' \emph{arXiv preprint arXiv:2212.06817}, 2022.

\bibitem{brohan2023rt}
A.~Brohan, N.~Brown, J.~Carbajal, Y.~Chebotar, X.~Chen, K.~Choromanski, \emph{et~al.}, ``Rt-2: Vision-language-action models transfer web knowledge to robotic control,'' \emph{arXiv preprint arXiv:2307.15818}, 2023.

\bibitem{o2023open}
A.~O'Neill, A.~Rehman, A.~Gupta, A.~Maddukuri, A.~Gupta, A.~Padalkar, A.~Lee, \emph{et~al.}, ``Open x-embodiment: Robotic learning datasets and rt-x models,'' \emph{arXiv preprint arXiv:2310.08864}, 2023.

\bibitem{Zhou_2023_ICCV}
B.~Zhou, H.~Zhou, T.~Liang, Q.~Yu, S.~Zhao, Y.~Zeng, J.~Lv, S.~Luo, Q.~Wang, X.~Yu, H.~Chen, C.~Lu, and L.~Shao, ``Clothesnet: An information-rich 3d garment model repository with simulated clothes environment,'' in \emph{ICCV}, October 2023, pp. 20\,428--20\,438.

\bibitem{10341573}
X.~Yu, S.~Zhao, S.~Luo, G.~Yang, and L.~Shao, ``Diffclothai: Differentiable cloth simulation with intersection-free frictional contact and differentiable two-way coupling with articulated rigid bodies,'' in \emph{IROS}, 2023, pp. 400--407.

\bibitem{sohn2015learning}
K.~Sohn, H.~Lee, and X.~Yan, ``Learning structured output representation using deep conditional generative models,'' \emph{NIPS}, 2015.

\bibitem{qi2017pointnet++}
C.~R. Qi, L.~Yi, H.~Su, and L.~J. Guibas, ``Pointnet++: Deep hierarchical feature learning on point sets in a metric space,'' \emph{Advances in neural information processing systems}, vol.~30, 2017.

\bibitem{touvron2023llamaopenefficientfoundation}
H.~Touvron, T.~Lavril, G.~Izacard, X.~Martinet, \emph{et~al.}, ``Llama: Open and efficient foundation language models,'' \emph{arXiv preprint arXiv:2302.13971}, 2023.

\bibitem{ravi2024sam}
N.~Ravi, V.~Gabeur, Y.-T. Hu, R.~Hu, C.~Ryali, T.~Ma, H.~Khedr, R.~R{\"a}dle, C.~Rolland, L.~Gustafson, \emph{et~al.}, ``Sam 2: Segment anything in images and videos,'' \emph{arXiv preprint arXiv:2408.00714}, 2024.

\bibitem{liang2018gpu}
J.~Liang, V.~Makoviychuk, A.~Handa, N.~Chentanez, M.~Macklin, and D.~Fox, ``Gpu-accelerated robotic simulation for distributed reinforcement learning,'' in \emph{Conference on Robot Learning}.\hskip 1em plus 0.5em minus 0.4em\relax PMLR, 2018.

\bibitem{li2018learning}
Y.~Li, J.~Wu, R.~Tedrake, J.~B. Tenenbaum, and A.~Torralba, ``Learning particle dynamics for manipulating rigid bodies, deformable objects, and fluids,'' \emph{arXiv preprint arXiv:1810.01566}, 2018.

\bibitem{bertiche2020cloth3d}
H.~Bertiche, M.~Madadi, and S.~Escalera, ``Cloth3d: clothed 3d humans,'' in \emph{ECCV}.\hskip 1em plus 0.5em minus 0.4em\relax Springer, 2020, pp. 344--359.

\bibitem{Ze2024DP3}
Y.~Ze, G.~Zhang, K.~Zhang, C.~Hu, M.~Wang, and H.~Xu, ``3d diffusion policy: Generalizable visuomotor policy learning via simple 3d representations,'' in \emph{RSS}, 2024.

\bibitem{ravi2024sam2}
N.~Ravi, V.~Gabeur, \emph{et~al.}, ``Sam 2: Segment anything in images and videos,'' \emph{arXiv preprint arXiv:2408.00714}, 2024.

\end{thebibliography}

\end{document}